\pdfoutput=1

\documentclass[11pt]{article}

\usepackage{acl}

\usepackage{times}
\usepackage{latexsym}

\usepackage[T1]{fontenc}

\usepackage[utf8]{inputenc}

\usepackage{microtype}

\usepackage{enumitem}
\usepackage{amsmath,amssymb,amsfonts}
\usepackage{bbm,dsfont}
\usepackage{mathrsfs}
\usepackage{graphicx}
\usepackage{multirow}
\usepackage{subfigure}
\usepackage{xspace}
\usepackage{booktabs}
\usepackage{comment}
\usepackage[normalem]{ulem}
\usepackage[misc]{ifsym}

\newcommand{\ourapproach}{\textsc{CompCoder}\xspace}

\definecolor{applegreen}{rgb}{0.55, 0.71, 0.0}

\definecolor{hycolor}{rgb}{0.7,0.7,0.3}

\makeatletter
\def\thanks#1{\protected@xdef\@thanks{\@thanks
        \protect\footnotetext{#1}}}
\makeatother

\title{Compilable Neural Code Generation with Compiler Feedback}

\author{
Xin Wang$^1$\textsuperscript{$\star\diamond$}\thanks{$\star$\ Equal contribution.}\thanks{$\diamond$ Work done while this author was an intern at Huawei Noah's Ark Lab.},
Yasheng Wang$^2$\textsuperscript{$\star$},
Yao Wan$^4$,
Fei Mi$^2$,
Yitong Li$^{2,3}$,
\\
{\bf 
Pingyi Zhou$^2$, 
Jin Liu$^1$\textsuperscript{\Letter}\thanks{\Letter\ Correspondence author.},
Hao Wu$^5$,
Xin Jiang$^2$,
Qun Liu$^2$}
\\
$^1$School of Computer Science, Wuhan University, China\\
$^2$Huawei Noah’s Ark Lab, \quad
$^3$Huawei Technologies Co., Ltd.\\
$^4$School of Computer Sci. \& Tech., Huazhong University of Science and Technology, China\\
$^5$School of Information Science and Engineering, Yunnan University, China
\\
\text{\{xinwang0920, jinliu\}@whu.edu.cn}, \
\text{wanyao@hust.edu.cn}, \
\text{haowu@ynu.edu.cn}\\
\text{\{wangyasheng, feimi2, liyitong3, zhoupingyi, Jiang.Xin, qun.liu\}@huawei.com}
}

\begin{document}
	\maketitle
	
	\begin{abstract}
        Automatically generating compilable programs with (or without) natural language descriptions has always been a touchstone problem for computational linguistics and automated software engineering.
        Existing deep-learning approaches model code generation as text generation, either constrained by grammar structures in decoder, or driven by pre-trained language models on large-scale code corpus (e.g., CodeGPT, PLBART, and CodeT5).
        However, few of them account for compilability of the generated programs.
		To improve compilability of the generated programs, 
        this paper proposes \ourapproach, a three-stage pipeline utilizing compiler feedback for compilable code generation, including language model fine-tuning, compilability reinforcement, and compilability discrimination. 
        Comprehensive experiments on two code generation tasks demonstrate the effectiveness of our proposed approach, improving the success rate of compilation from 44.18 to 89.18 in code completion on average and from 70.3 to 96.2 in text-to-code generation, respectively, when comparing with the state-of-the-art CodeGPT.
	\end{abstract}
	
	\section{Introduction}
Automated code generation (or program synthesis) has attracted much
	attention over the past few years~\cite{lu2021codexglue}, because of its potential to improve the productivity of developers, as well as to speed up the software development~\cite{parvez2021retrieval, wang2021syncobert}. 
	In the life cycle of software development, different types of code generation tasks are desired, including code completion~\cite{liu2020multi, liu2020a}, text-to-code generation~\cite{hashimoto2018a}, program translation~\cite{Chen2018TreetotreeNN}, and program repair~\cite{Yasunaga2021BreakItFixItUL}.

	Recently, much effort has been made
	to advance the development of code generation~\cite{li2018code}, using different logical forms of code, such as the abstract syntax tree (AST)~\cite{kim2021code, yin2017a, rabinovich2017abstract}, sketch~\cite{Nye2019LearningTI} and graph~\cite{Yasunaga2020GraphbasedSP}.
	Benefiting from the strong power of pre-training techniques~\cite{devlin2019bert, Wang2021ServiceBERTAP} in natural language processing, 
	several attempts have been made towards pre-training a language model on large-scale code corpus for code generation, such as CodeGPT~\cite{lu2021codexglue},  PLBART~\cite{ahmad2021unified}, and CodeT5~\cite{wang2021codet5}.
	
	\begin{figure}
		\centering
		\includegraphics[width=0.5\textwidth]{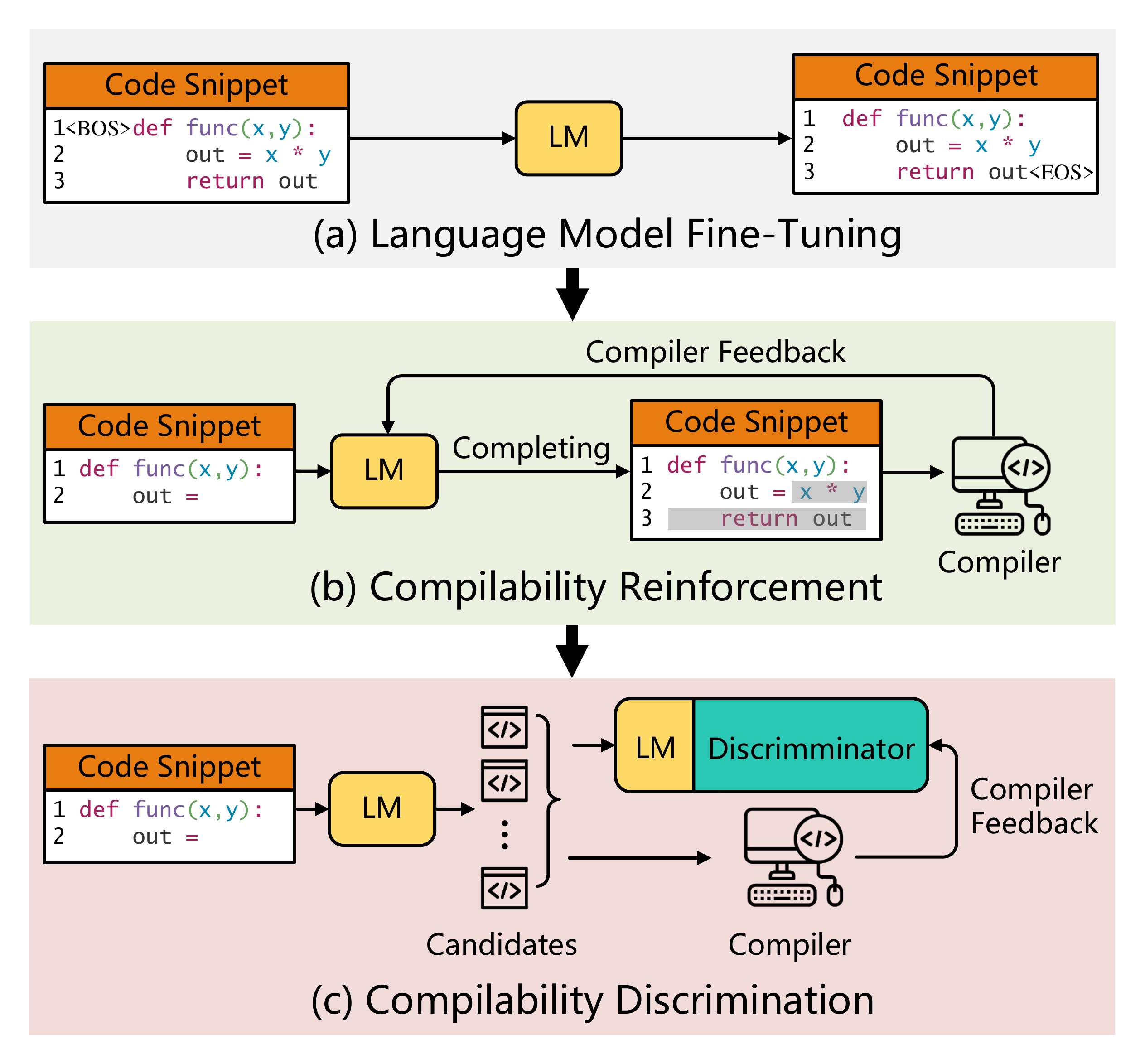}
		\caption{\label{fig:abs} 
			 An illustration of Python code completion by \ourapproach, utilizing the compiler feedback with three stages.
		}
	\end{figure}

    However, to the best of our knowledge, 
    most deep-learning approaches for code generation are still difficult to guarantee the compilability of the generated code, resulting in non-compilable code.
    For example, \citet{Chen2021SequenceRSL} found that up to 67\%-97\% of patches generated by the most advanced deep-learning-based models are non-compilable. 
    We think this is because they generally do not directly optimize the compilability for code generation.
    The generation of non-compilable code will waste the time of programmers, as well as seriously reduce the trust and satisfaction of developers with the model. 
    To improve the compilability of the generated code, some works attempt to repair the synthesized program which fails to compile~\cite{Kulal2019SPoCSP, Yasunaga2020GraphbasedSP, Yasunaga2021BreakItFixItUL}. Recently, 
    \citet{Korbak2021EnergyBasedMF} attempt to directly generate compilable code using an energy model with compilability constraints.

	This paper focuses on the task of compilable neural code generation. Different from previous works, we use compilability signals in two ways and design a novel method to jointly train the discriminator and generator for compilable code generation. 
	Concretely, we propose \ourapproach, a novel three-stage pipeline utilizing compiler feedback for compilable code generation, including language model fine-tuning, compilability reinforcement, and compilability discrimination. 
    Figure~\ref{fig:abs} shows an example of Python code completion by \ourapproach, which utilizes the compiler feedback in two ways.
    In Figure~\ref{fig:abs}(b), we use the compiler feedback to optimize the generator. In Figure~\ref{fig:abs}(c), we use the discriminator to check if the results generated by the generator can be successfully compiled. The joint training of the generator and discriminator significantly improves the compilability of the generated code. 

	Overall, the key contributions of this paper are as follows:

	\begin{itemize}[itemsep=-1pt,topsep=0pt,leftmargin=*]
		\item 
		We use compilability signals in two ways and design a novel method to jointly train the generator and discriminator for compilable code generation, called \ourapproach. 
        We refine a pre-trained code generator using reinforcement learning and jointly learn a discriminator to enforce the generator to correct its own mistakes.
		\item 
        Comprehensive experiments on two code generation tasks demonstrate the effectiveness of \ourapproach. It boosts the average compilation rate of CodeGPT from 44.18 to 89.18 in the code completion task and from 70.3 to 96.2 in the text-to-code generation task.
	\end{itemize}
	\begin{figure*}
		\centering
		\includegraphics[width=0.98\textwidth]{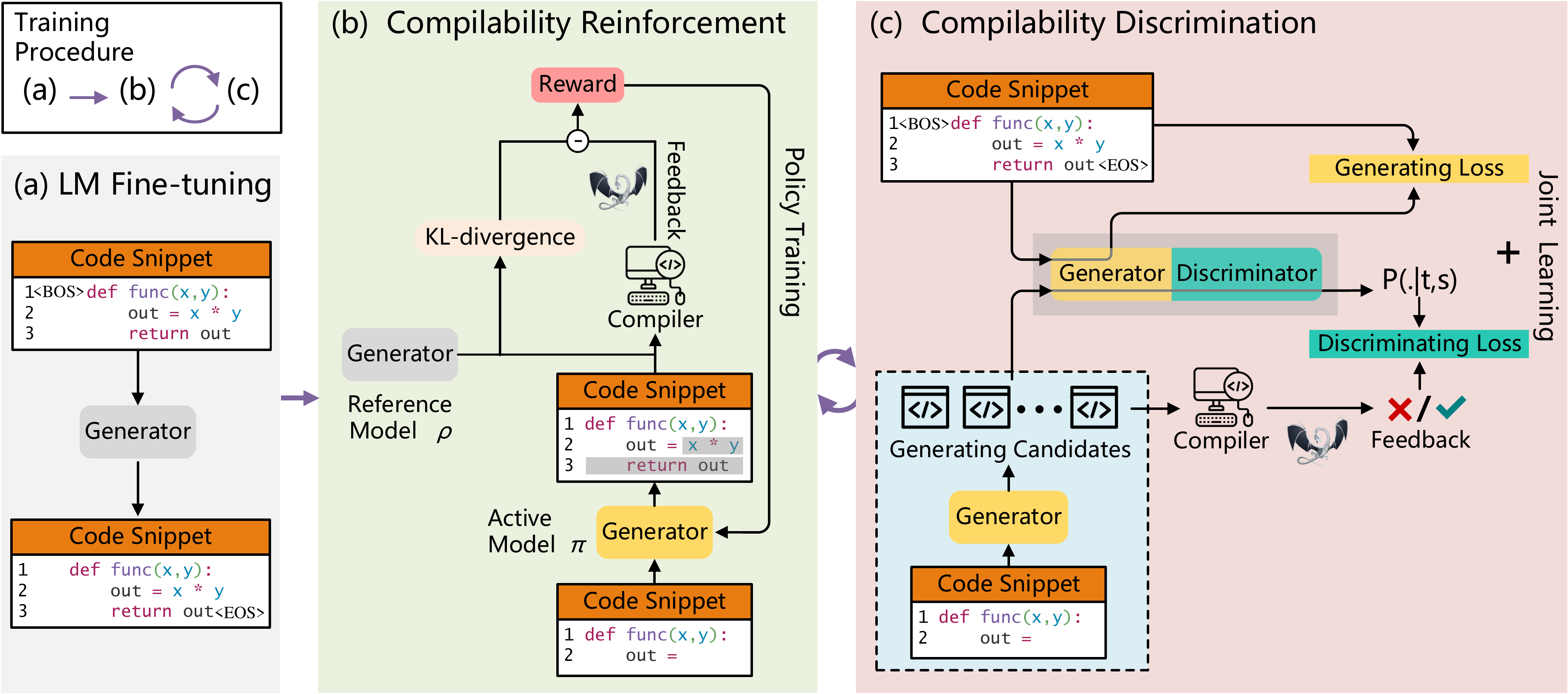}
		\caption{\label{fig:model} 
			 An illustration of our proposed three-stage pipeline for Python code completion. (a) We first fine-tune the generator based on pre-trained language models. (b) 
			 We take the compiler feedback into account as a reward via reinforcement learning.
			 (c) 
			 We design a compilability discriminator which is jointly trained with the generator, to enforce the generator to correct its own mistakes.
			 Stages 2 and 3 are performed alternately.
		}
	\end{figure*}

	\section{Preliminary}
	In this section, we set out notations for task formulation, as well as some preliminaries of compiler feedback.
	Let $s \in \mathcal{S}$ denote a given input, which can be a piece of partial code, natural-language description, or buggy program. Let $t \in \mathcal{T}$ denote the generated source code.
	Formally, the problem of code generation can be formulated as learning a mapping $f$ between the input space and target code space, i.e. $f:\mathcal{S}\rightarrow \mathcal{T}$.
	In this paper, we investigate two specific code generation tasks, code completion and text-to-code generation, conditioned on different inputs.
    
    \paragraph{Code Completion} Let $c = \{c_1,c_2, \ldots, c_{|c|}\}$ denote a sequence of code tokens for program $c$, where $|c|$ denotes the length of the code.
	We use notation $c_{1 \colon m} \in \mathcal{S}$ to refer to the previous code snippet $\{c_1,c_2,\ldots,c_m\}$ and notation $c_{m+1 \colon |c|} \in \mathcal{T}$ 
	to represent the subsequent code snippet $ \{c_{m+1},\ldots,c_{|c|}\}$. 
	The code completion task can be defined as generating the subsequent ($t$) code token sequence $c_{m+1 \colon |c|}$, given the previous ($s$) code sequence $c_{1 \colon m}$.
	\paragraph{Text-to-Code Generation}
	Different from code completion, text-to-code generation aims to generate a whole program based on natural language description.
	Let $d = \{d_1, d_2, \ldots , d_{|d|}\}$ refer to a sequence of natural-language tokens. The text-to-code generation task can be defined as generating source code $c=t \in \mathcal{T}$, given the corresponding natural language description $d=s \in \mathcal{S}$. 
	\paragraph{Compiler Feedback}
    As the whole program $c$ is generated, no matter from partial code snippets or natural-language descriptions, we feed it into a compiler to test whether it can be compiled successfully.
    Formally, we define the the compiler feedback as:
	\begin{equation}
    \label{eq:feedback}
        {\rm \textit{feedback}} = {\mathds{1}_{\operatorname{Compiler}}}(c) \,,  
    \end{equation}
 where the compiler feedback is a binary value (compilable or non-compilable), and $c$ denotes the code snippet fed into the compiler. 
 As for the task of text-to-code generation, we simply feed the generated code $t$ into the compiler, i.e., $c=t$. 
 As for the task of code completion, we concatenate the partial code with generated code as a whole program, i.e., $c=[s;t]$, where $;$ is the concatenation operation.
 
	\section{\ourapproach}
		\label{sec:overview}
	Figure~\ref{fig:model} shows the overall architecture of \ourapproach on the code completion task, which covers three stages, i.e., language model fine-tuning (Stage 1), compilability reinforcement  
	(Stage 2) and compilability discrimination (Stage 3). 
	In the following subsections, we will elaborate on each stage one by one.
	We alternately perform Stages 2 and 3, as described in Section~\ref{pipeline}.

   	\subsection{Stage 1: Language Model Fine-Tuning}
	As shown in Figure~\ref{fig:model}(a), we adopt CodeGPT as the generator, which uses GPT-2~\cite{Radford2019LanguageMA} as the starting point and is continually pre-trained on the large-scale code corpus.
    Our generator is then fine-tuned on the target task to minimize the cross-entropy loss:
	\begin{equation}
		\label{eq:gen}
		\mathcal{L}_{\rm G} = -\frac{1}{|\mathcal{M}|} \sum^{|\mathcal{M}|}_i \sum^{|\mathcal{V}|}_j \ Y_{ij}\ {\rm log}\ P_{ij}\,,
	\end{equation}
	where $\mathcal{M}$ denotes the set of the generated code tokens, 
	$\mathcal{V}$ represents the vocabulary, $Y_{ij}$ denotes the label of the code token $i$ in class $j$, and $P_{ij}$ is the predicted probability of token $i$ in class $j$.
	
	During training, the generator takes $x= \{ \texttt{<BOS>}, c, \texttt{<EOS>}\}$ as the input in the code completion task, and $x= \{ d, \texttt{<BOS>}, c, \texttt{<EOS>}\}$ as input in the text-to-code generation task, correspondingly.
	Special tokens $\texttt{<BOS>}$ and $\texttt{<EOS>}$ indicate the start and end symbols of code sequences.
	After several epochs of supervised fine-tuning on the target task dataset, we save the trained generator, which will be used in the next stage.
	
	\begin{figure*}
		\centering
		\includegraphics[width=0.84\textwidth]{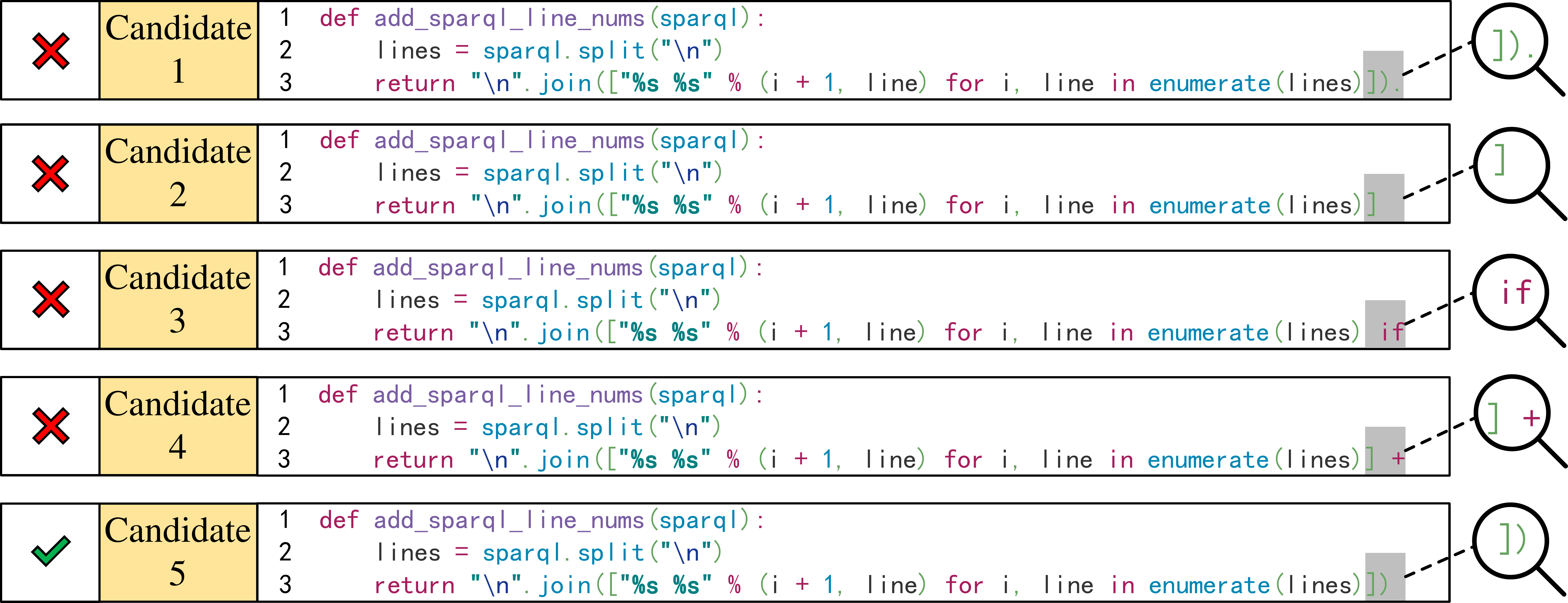}
		\caption{\label{fig:rank}
			An example of code completion. We mask the last five tokens of the code and let the generator complete them. Some minor mistakes prevent four candidates from being correctly compiled by the program compiler.
		}
	\end{figure*}
    \subsection{Stage 2: Compilability Reinforcement}
	Reinforcement Learning (RL) is a method of learning the optimal policy by obtaining reward signals from the real environment~\cite{Sutton1998IntroductionTR, wan2018improving}. As shown in Figure \ref{fig:model}(b), we use the fine-tuned generator $\rho$ (after Stage 1) as the \textit{reference model}. Then we initialize a policy $\pi=\rho$.
	Given an input sequence $s \in \mathcal{S}$, our goal is to find a policy $\pi$ that generates an output sequence $t \in \mathcal{T}$ with the objective of maximizing the compilability-based reward. We use RL (specifically PPO2 version of Proximal Policy Optimization~\cite{Schulman2017ProximalPO}) to directly optimize the expected reward as:
	\begin{equation}
		\mathbb{E}_{\pi} \left[r\right] = \mathbb{E}_{s\sim \mathcal{S}, t\sim \pi(.|s)}\left[r(s, t)\right], 
	\end{equation} 
	where the policy $\pi$ is rewarded by the compiler (Eq.~\ref{eq:feedback}), $r$ is the reward function. 
	We define $r(s,t) = 1.0$ iff the code can be compiled by the program compiler and $r(s,t) = -1.0$ otherwise.
	
	It is worth mentioning that code compilability constraints can be strong or weak. \textit{Strong constraint} is defined that a long piece of code snippet may not be correctly compiled if a certain token is changed. And \textit{weak constraint} means a blank string consisting of whitespace characters can be correctly compiled by the compiler. Concretely, in the text-to-code generation task, if the generator generates a string composed of whitespace characters, the compiler will consider it as a good case. In the code completion task, if the previous code snippet is compilable, the generator can fool the compiler easily. The RL is good at making use of this, resulting in the generated code can be compiled, but seriously deviating from the generation likelihood objective.

	To avoid \textit{active model} $\pi$ being too far away from  \textit{reference model} $\rho$, we add a Kullback-Leibler (KL) penalty with expectation, e.g., $\beta$KL$(\pi,\rho)$~\cite{Ziegler2019FineTuningLM}. 
    Therefore, the modified reward will be reformulated as follows:
	\begin{equation}
		r(s,t) = r(s,t) - \beta\ {\rm log}\frac{\pi (t|s)}{\rho (t|s)}\,,
	\end{equation}
	where $\beta$ is a constant, which plays the role of an entropy bonus, preventing the policy from moving too far from the range where $r$ is valid.
	
	To alleviate the imbalance between the reward term and the KL penalty term and improve the stability of training, we use autoregressive fine-tuning (Causal Language Modeling)~\cite{Radford2019LanguageMA} to make the KL penalty term fluctuate within a small range after RL training. This fine-tuning process incorporates a compilability-aware discriminator that will be introduced in the next stage.

    \subsection{Stage 3: Compilability Discrimination}
     Figure~\ref{fig:rank} shows an example of code completion. We mask the last five tokens of a Python function and ask the generator to complete them. The generator generates five candidates with high probabilities.
	Some minor mistakes prevent four of them from being successfully compiled. We hope the generator can have more perception power to explicitly distinguish compilable and non-compilable code generated by itself. Therefore, at this stage, we design a compilability-aware discriminator to deal with this issue.
    
	Concretely, we add a discriminator (a two-layer MLP equipped with the ${\rm tanh}$ activation function between layers) after the final hidden layer of the generator. 
	As shown in Figure~\ref{fig:model}(c), given the input sequence ($s$), we perform beam search on the generator to generate top-$k$ candidates ($t$).
	Each entire code $c \in \mathcal{Q}$ ($c=[s;t]$ in the code completion task) is labeled by the program compiler as positive (1) or negative (0), depending on whether it can be successfully compiled (see Eq.~\ref{eq:feedback}).
	
	We use the hidden representation of the last token ($\texttt{<EOS>}$) as the final representation of the entire code $c$.
	Finally, the hidden representation of the last token ($\texttt{<EOS>}$) is fed into the discriminator for prediction:
	\begin{eqnarray}
		h_{\rm \texttt{<EOS>}} &=& {\rm CodeGPT}(s,t)\,,\\
		h^{'}_{\rm \texttt{<EOS>}} &=& {\rm Discriminator}(h_{\rm \texttt{<EOS>}})\,,\\
		P(.|t,s) &=&  {\rm softmax}(h^{'}_{\rm \texttt{<EOS>}})\,,
	\end{eqnarray}
	where $h_{\rm \texttt{<EOS>}}$ denotes the representation of the last token $\texttt{<EOS>}$. 
	The training loss of the discrimination process can be defined as: 
    \begin{equation}
        \begin{aligned}
        \mathcal{L}_{\rm D} &=- \frac{1}{|\mathcal{Q}^{+}\cup \mathcal{Q}^{-}|} \left[ \sum_{c\in \mathcal{Q}^{+}} \log\ P{(1|t,s)} \right.
        \\
        &\left.{+  \sum_{c\in \mathcal{Q}^{-}} \log\ P{(0|t,s)}}\right]\,,
        \end{aligned}
    \end{equation}
	where $\mathcal{Q}^{+}$ and $\mathcal{Q}^{-}$ represent positive and negative sets respectively. The parameters of the generator and discriminator will be jointly updated.
	
	At this stage, we jointly train the generator and discriminator, including a generating objective (\textit{to learn the generator only}) and a discriminating objective (\textit{to learn the generator and discriminator together}), as shown in Figure~\ref{fig:model}(c). 
	The joint training loss is defined as follows:
	\begin{equation}
		\mathcal{L} = \mathcal{L}_{\rm G} + \mathcal{L}_{\rm D}\,.
	\end{equation}
	
	\subsection{Overall Pipeline}
	\label{pipeline}
	\paragraph{Training Procedure} We perform an interactive training procedure. Concretely, except that the first epoch contains Stages 1, 2, and 3, each subsequent epoch only consists of Stages 2 and 3. We update the \textit{reference model} (at Stage 2), and \textit{candidates} in Stage 3 is generated on the training dataset, which is time consuming, so we update the \textit{candidates} in a preset frequency.
	
	For better understanding, Stage 2 improves the compilability of generated code, Stage 3 distinguishes the compilable and non-compilable code generated by itself. Stage 2 and 3 refine each other and improve the performance iteratively, which is a basic idea of this training procedure.
	We think that the generator with high compilability (after Stage 2) facilitates the learning of the discriminator (discriminating objective at Stage 3).
	The autoregressive fine-tuning (generating objective at Stage 3) helps the KL penalty term (at Stage 2) fluctuate in a small range, improving the stability of RL training. 
	At Stage 3, the discriminating objective is optimized by learning the generator and discriminator together, which makes the generator have more perception power to distinguish compilable and non-compilable code.
	\paragraph{Inference Procedure}
	The model inference consists of two stages. Given an input sequence $(s)$, we perform the beam search on the generator to generate top-$k$ candidates.
	The code ($c$ in Eq.~\ref{eq:feedback}) with the highest compilability probability evaluated by the discriminator will be selected. Then the output ($t$) can be obtained as the final result.

	\section{Experiment Setup}
	\subsection{Evaluation Tasks and Datasets}
	We conduct experiments on two tasks: code completion and text-to-code generation. To investigate the compilability of the generated code, we need to preserve the indentation and newline operations in code. We also need to make sure that the code and its version belong to the scope of the compiler.
	Existing datasets on both of the two tasks usually do not serve these considerations.
	For convenience, we choose Python for experiments, as it is very popular and used in many projects. We conduct all experiments based on Python 3 environment and adopt the \texttt{codeop}\footnote{\url{https://docs.python.org/3.6/library/codeop.html}} module to simulate the program compiler. We remove code that could not be compiled correctly by the compiler.
	
	\paragraph{Code Completion}
	For the code completion task, we use the Python corpus in CodeSearchNet~\cite{Husain2019CodeSearchNetCE}. We want to study the compilability of long enough code, while longer code means higher computational overhead. Therefore, we extract 50$k$ compilable Python methods (Python 3 version) with eclectic token lengths ranging from 64 to 96. We randomly select 45$k$ samples for training and the remaining 5$k$ samples for testing. We mask a different number of tokens at the tail of the source code and let the model complete.
	
	\paragraph{Text-to-Code Generation}
	For the text-to-code generation task, we adopt the AdvTest dataset~\cite{lu2021codexglue}, which contains 251,820 text and Python code pairs. We only need code in Python 3 version. We expect code token lengths to range from 128 to 170, a moderate length, and text token lengths to be at least more than 5, containing sufficient semantics. Finally, we extract about 41$k$ text-code pairs. We randomly select 40$k$ text-code pairs for training, and the remaining 1$k$ text-code pairs for testing. 
	
	\subsection{Evaluation Metrics}
	To evaluate the quality of the generated code, we adopt two widely-used evaluation metrics: Levenshtein \textit{Edit Similarity (ES)}~\cite{svyatkovskiy2020intellicode, lu2021codexglue} and \textit{Compilation Rate (CR)}~\cite{Kulal2019SPoCSP}. Levenshtein Edit Similarity measures the number of single-character edits required to transform one string into another. It is a critical evaluation metric for the code generation scenario, as it measures the effort required for the developer to correct the code. Compilation Rate measures how many code can be correctly compiled by the program compiler. For both of these metrics, bigger values indicate better performance.
	
		\begin{figure*}[t!]
		\centering
		\subfigure[]{\label{fig:cc_b} \includegraphics[width=3.83cm]{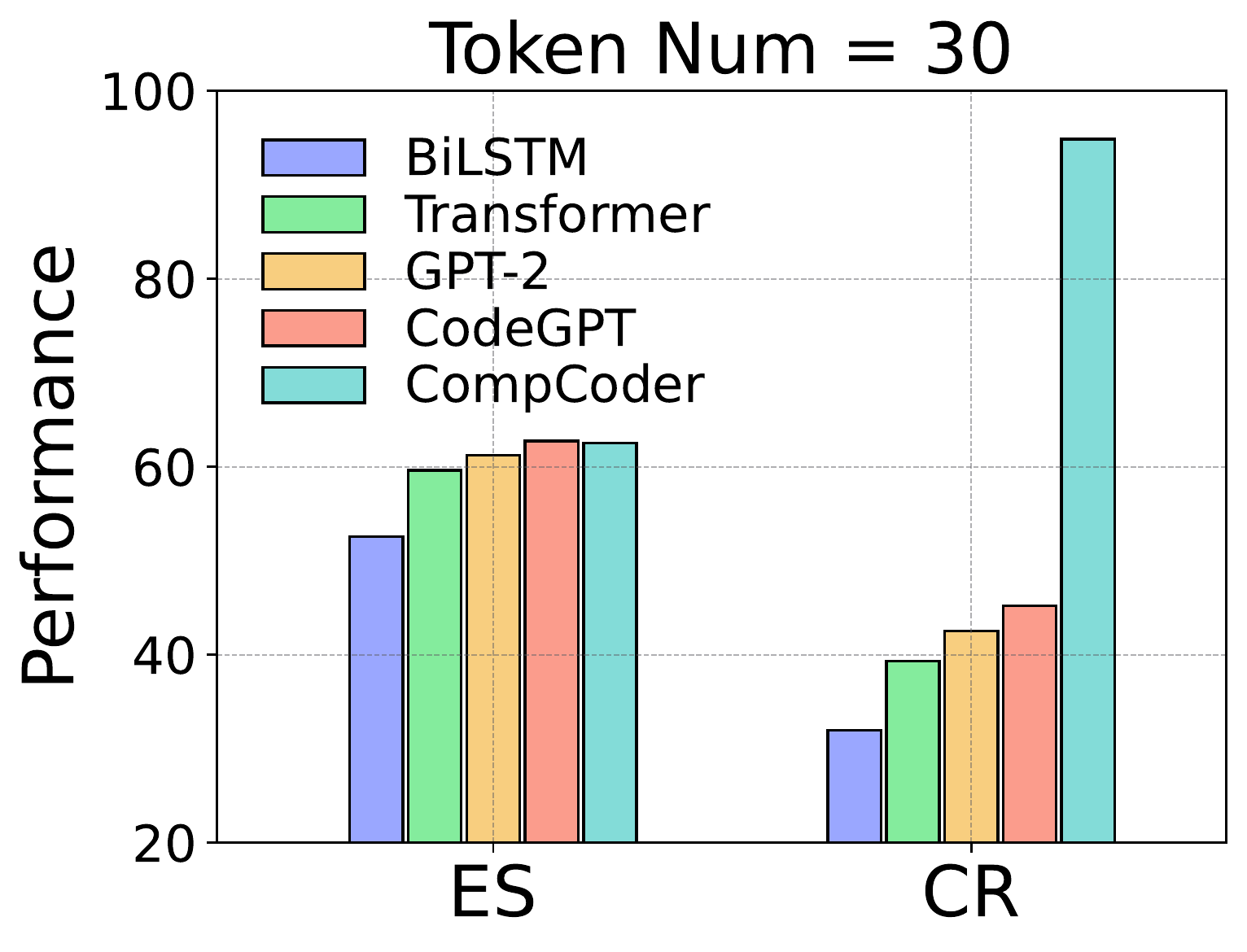}}
		\subfigure[]{\label{fig:cc_c} \includegraphics[width=3.83cm]{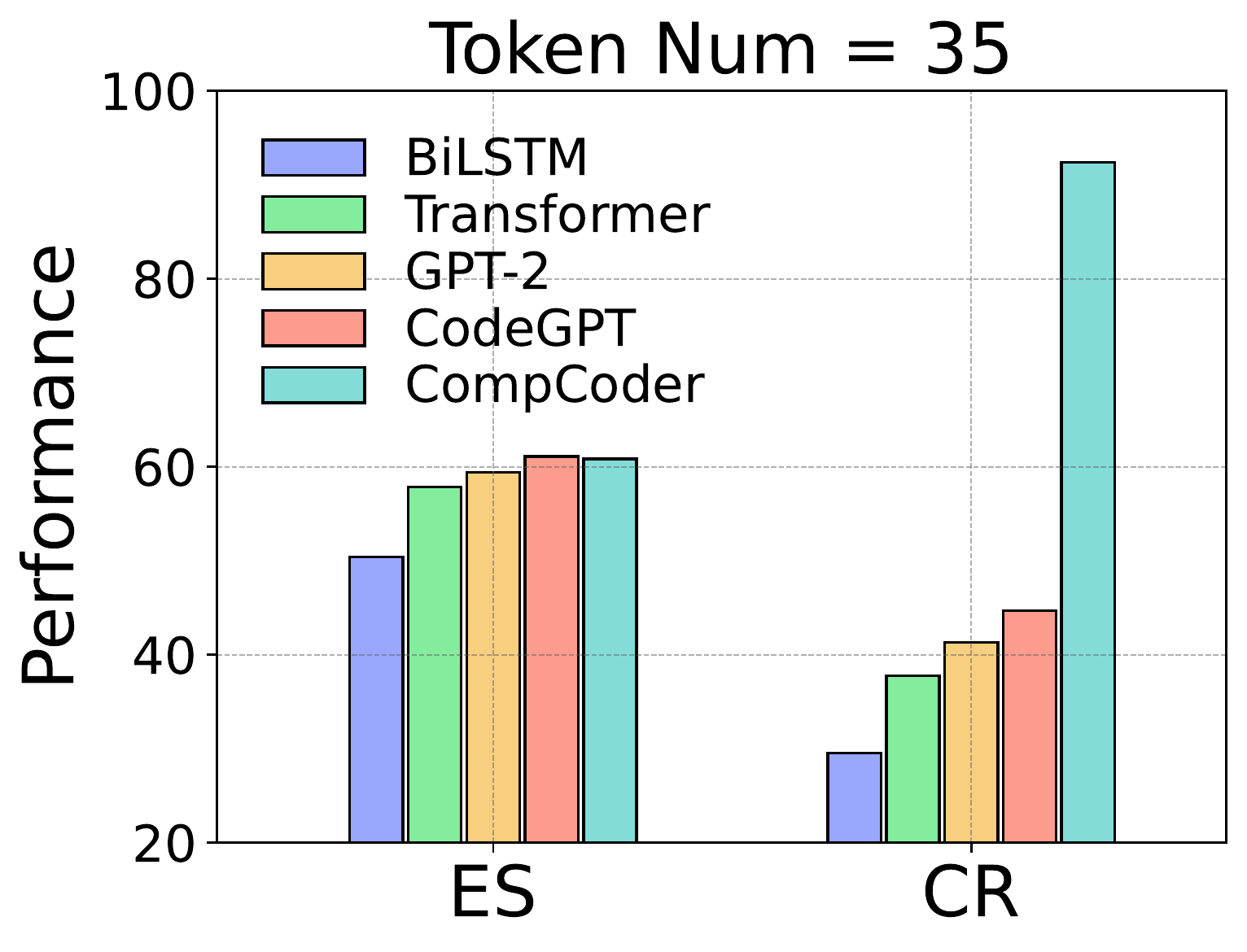}}
		\subfigure[]{\label{fig:cc_d} \includegraphics[width=3.83cm]{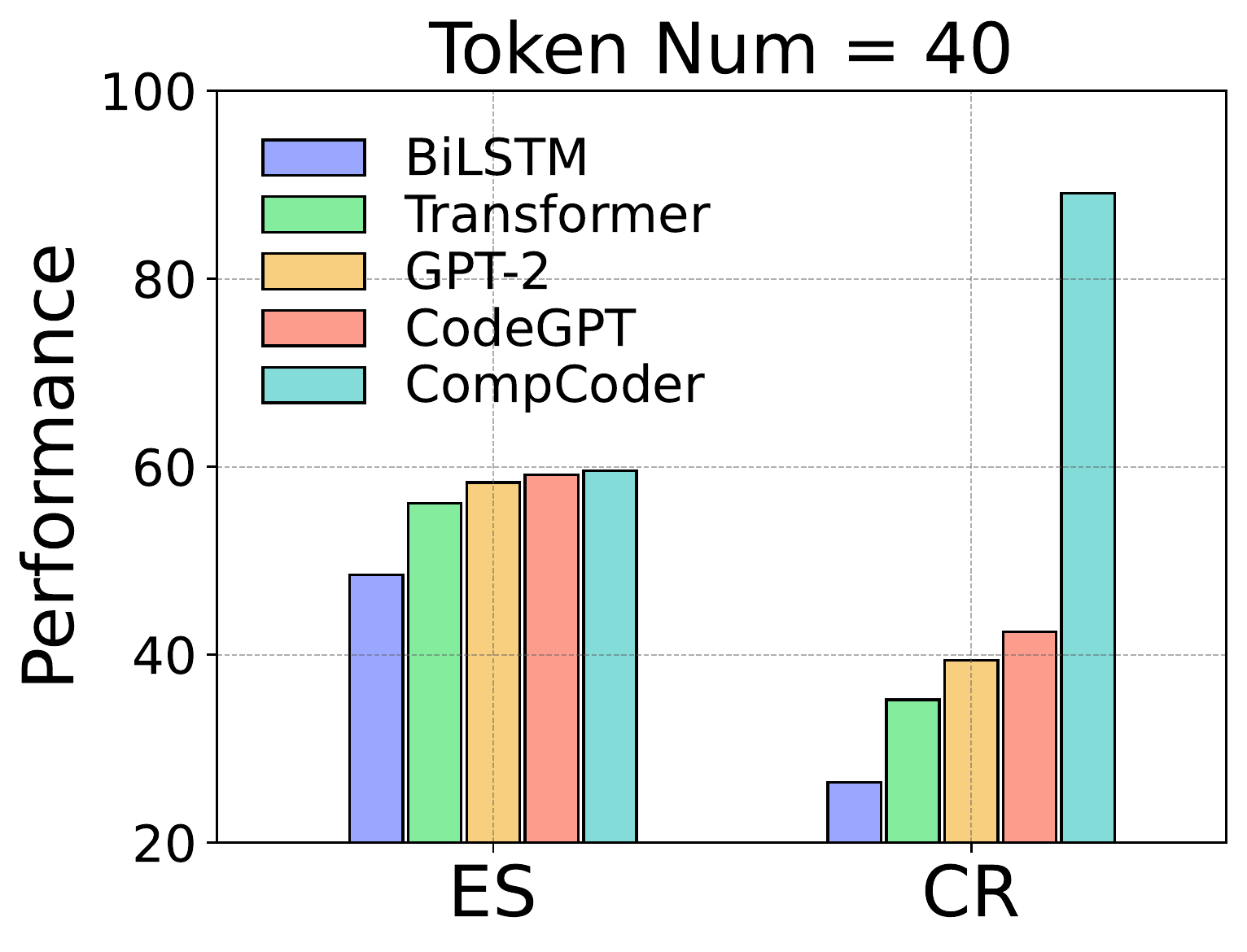}}
		\subfigure[]{\label{fig:cc_e} \includegraphics[width=3.83cm]{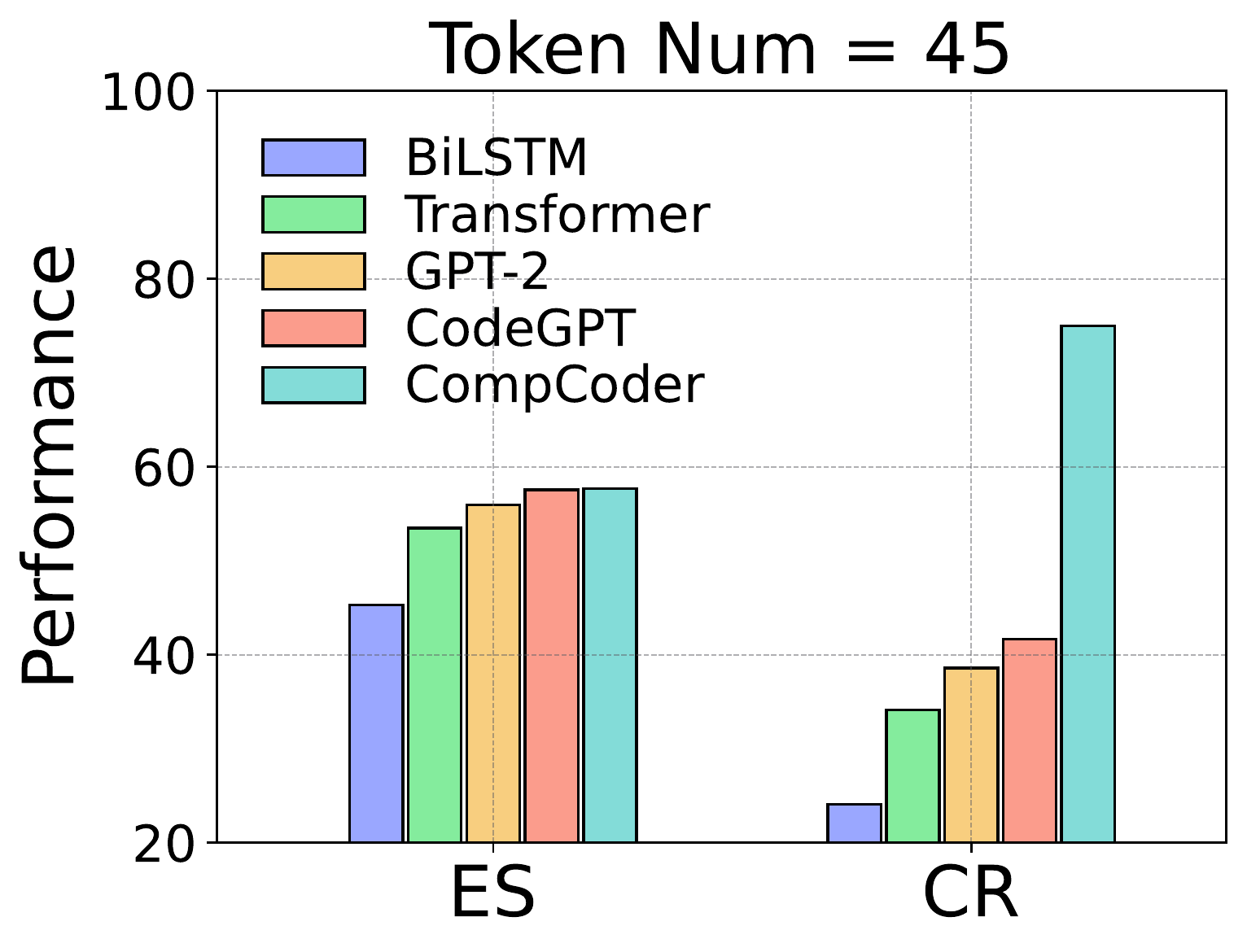}}
		\caption{\label{fig:cc} : Results in the code completion task (completing 30, 35, 40, 45 tokens respectively) evaluating with Edit Similarity (ES) and Compilation Rate (CR) metrics, using the CodeSearchNet-Python dataset.}
	\end{figure*}
	\subsection{Baseline Methods}
	We compare our approach with various state-of-the-art models in the code completion task and the text-to-code generation task:

	\begin{itemize}[itemsep=-1pt,topsep=0pt,leftmargin=*]
		\item \textbf{BiLSTM} is a Seq2Seq model based on Bidirectional LSTM with an attention mechanism~\cite{Luong2015EffectiveAT}.
		
		\item \textbf{Transformer}~\cite{vaswani2017attention} is the base architecture of CodeGPT. We use 6-layer Transformer decoder to conduct experiments.
		
		\item \textbf{GPT-2}~\cite{Radford2019LanguageMA} is an auto-regressive pre-trained model trained on large-scale text corpus.
		
		\item \textbf{CodeGPT}~\cite{lu2021codexglue} is pre-trained with source code corpus on the basis of GPT-2 vis causal language modeling objective~\cite{Radford2019LanguageMA}.
		
		\item \textbf{PLBART}~\cite{ahmad2021unified} is based on the BART~\cite{lewis2020bart} architecture, which is pre-trained on large-scale Java and Python corpora via denoising autoencoding.
		
	    \item \textbf{CodeT5}~\cite{wang2021codet5} is based on the T5~\cite{Raffel2020ExploringTL} architecture, which employs denoising sequence-to-sequence pre-training on multiple programming languages.
	\end{itemize}

	\subsection{Implementation Details}
	In the code completion task, we set the learning rate as $1.5e\text{-}5$, the batch size as 32, the maximum fine-tuning epoch as 20, the maximum code sequence length as 96. 
	We mask different numbers of code tokens (25, 30, 35, 40, and 45) and ask the model to complete them. We set the minimum generation length as 25, 30, 35, 40, and 45 accordingly.
	In the text-to-code generation task, we set the learning rate as $1.5e\text{-}5$, the batch size as 16, the maximum fine-tuning epoch as 20, the maximum text and code sequence length as 32 and 170. We set the minimum generation length as 96 (the generated code is slightly shorter than the ground-truth is allowed). In these two tasks, the generated sequence consisting of whitespace characters will be considered as a bad case.
	
	We use the Adam optimizer to update model parameters. We train our model on the basis of CodeGPT checkpoint\footnote{\url{https://huggingface.co/microsoft/CodeGPT-small-py-adaptedGPT2}}. Our model is trained on 2 NVIDIA Tesla V100 with 32GB memory. We employ the same tokenizer as CodeGPT. 
	To train the policy $\pi$, we use the PPO2 version of Proximal Policy Optimization~\cite{Schulman2017ProximalPO}. In each epoch, we only randomly select 5\% training data for the stability of RL training (Stage 2). In other stages (Stages 1 and 3), we use the full training data. To generate \textit{candidates} (at Stage 3), we set the beam size as 5 in beam search. For efficiency, we update the \textit{candidates} every 5 epochs.

	\begin{table}[htbp]
		\normalsize
		\centering
		\setlength{\tabcolsep}{5mm}{
				\begin{tabular}{lcc}
						\toprule
						Models&ES& CR\\
						\midrule 
						BiLSTM&55.32&36.34\\
						Transformer&61.47&40.22\\
						GPT-2&63.02&43.26\\
						CodeGPT&64.47&46.84\\
						\ourapproach&\textbf{64.53}&\textbf{94.48}\\
						\bottomrule
				\end{tabular}
				\caption{Results in the code completion task (completing 25 tokens) evaluating with Edit Similarity (ES) and Compilation Rate (CR) metrics, using the CodeSearchNet-Python dataset.} 
				\label{table:codecompletion}
				}
	\end{table}
	
	\section{Results and Analysis}
	
	\subsection{Code Completion}
	Table~\ref{table:codecompletion} shows the results of the code completion task. We mask 25 tokens at the tail of code functions and ask the generation model to complete. We can observe that:
	(1) The code generated by existing autoregressive models has a low Compilation Rate. CodeGPT and GPT-2 only achieve 46.84 and 43.26 scores respectively on the Compilation Rate,  which means that more than half of the code generated by them cannot be correctly compiled by the program compiler.
	(2) \ourapproach significantly improves the Compilation Rate. It obtains 94.48 scores on the Compilation Rate, which is 47.64 points higher than the closest competitor (CodeGPT). 
	(3) When our approach significantly improves the Compilation Rate, it does not sacrifice the fluency of the generated code. \ourapproach obtains a comparable and even slightly better Edit Similarity score than other baselines, indicating that it effectively preserves the code fluency.

	Figure~\ref{fig:cc} presents more results of the code completion task in the setting of completing 30, 35, 40, and 45 tokens.
	\ourapproach still effectively improves the Compilation Rate when generating longer code.
    As the completion length increases, our approach outperforms CodeGPT by 49.66, 47.68, 46.64, and 33.36 points in the setting of completing 30, 35, 40, and 45 tokens, respectively. On average, our approach outperforms CodeGPT by \textit{45} points across a different number of tokens for the task of code completion.

	\begin{table}[htbp]
		\normalsize
		\centering
		\setlength{\tabcolsep}{5mm}{
		    \begin{tabular}{lcc}
				\toprule
				Models & ES & CR\\
				\midrule
				BiLSTM&54.86&48.7\\
				Transformer &57.47&55.6\\
				GPT-2 &60.54&63.3\\
				CodeGPT &61.82&70.3\\
				PLBART&62.43&71.9\\
				CodeT5&62.58&73.1\\
				\ourapproach& \textbf{62.74}&\textbf{96.2}\\
				\bottomrule
		    \end{tabular}
		    \caption{Results in the text-to-code generation task evaluating with Edit Similarity (ES) and Compilation Rate (CR), using the AdvTest dataset.} 
		    \label{table:codegen}
		    }
		
	\end{table}
	\subsection{Text-to-Code Generation}
	Table \ref{table:codegen} presents the results of the text-to-code generation task. 
	We could see that: (1) \ourapproach significantly outperforms all other models w.r.t. the Compilation Rate. E.g., \ourapproach achieves 23.1 points and 24.3 points improvements when compared with PLBART and CodeT5 respectively. (2) 
	Compared to code completion task (Table \ref{table:codecompletion}), all models in the text-to-code generation task have relatively higher Compilation Rate. One of the main reasons we think may be: code completion requires the generated code to be constrained by the existing (previous) code, which is a much stronger restriction than the text-to-code generation.
    
    \begin{table}[htbp]
		\normalsize
		\centering
		\setlength{\tabcolsep}{2mm}{
		    \begin{tabular}{lcc}
				\toprule
				Models & ES & CR \\
				\midrule
				(1) CodeGPT&64.47&46.84\\
				(2) \ \ w/ ${\rm \text{D}}_{\text{train}}$&65.46&64.88\\
				(3) \ \ w/ RL&64.71&76.48\\
				(4) \ \ w/ RL+${\text{D}}_{\text{train}}$&64.43&83.14\\
				(5) \ \ w/ ${\rm \text{D}}_{\text{train}}$+${\rm \text{D}}_{\text{test}}$ &65.24&81.96\\
				(6) \ \ w/ RL+${\rm \text{D}}_{\text{train}}$+${\rm \text{D}}_{\text{test}}$ (Ours) &64.53&94.48\\
				\bottomrule
				
		    \end{tabular}
		    \caption{Ablation study in the code completion task in the setting of completing 25 code tokens.}
		    \label{table:ablation}
		    } 
	\end{table}
	\subsection{Ablation Study}
	\label{section:ab}
	We compare several simplified versions
	of our model to understand contributions of different components, including the Reinforcement Learning (RL) component and the discriminator's effect for both model training (D$_{\text{train}}$) and model inference (D$_{\text{test}}$). As a case study, we take the code completion task as an example in the setting of completing 25 tokens and present the results in Table \ref{table:ablation}. 
    
    Several meaningful observations can be drawn: \textbf{First}, both RL (Row 2) and D$_{\text{train}}$ (Row 3) effectively increase the code Compilation Rate of the generation model (CodeGPT in Row 1),
    which confirms that the two components we designed can indeed improve the ability of the generator for compilable code generation. 
    \textbf{Second}, applying RL and D$_{\text{train}}$ together (Row 4) further improves the Compilation Rate over their individual contributions. \textbf{Third}, using the discriminator to select the output during model inference stage (D$_{\text{test}}$) is beneficial. It further boosts the Compilation Rate of vanilla ``D$_{\text{train}}$'' by 17.08\% (Row 5 v.s. Row 2) and boosts ``RL+${\text{D}}_{\text{train}}$'' by 11.34\% (Row 6 v.s. Row 4).
    \textbf{Forth}, these three components (RL, D$_{\text{train}}$, D$_{\text{test}}$) that effectively improve the Compilation Rate do not compromise the generation capability measured by the Edit Similarity.
    
    \begin{figure}
		\centering
		\includegraphics[width=7.5cm]{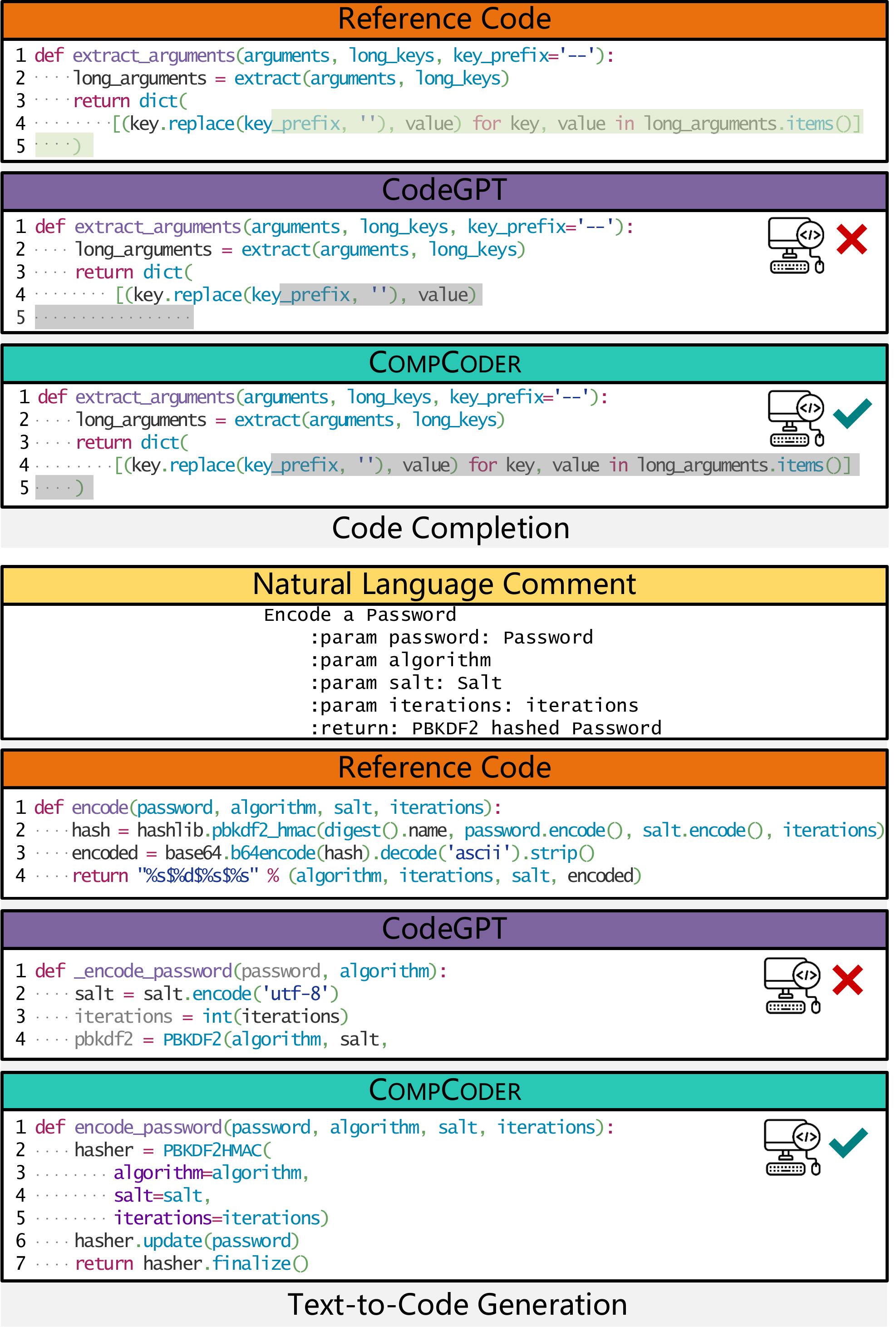}
		\caption{\label{fig:case} 
		    Case study for code completion and text-to-code generation tasks.
		}
	\end{figure}
    \subsection{Case Study}
    To better understand the effectiveness of our proposed approach, we present two cases for code completion and text-to-code generation tasks respectively.
    For both CodeGPT and \ourapproach, we present top-1 result in Figure~\ref{fig:case}.
    For code completion, we observe that CodeGPT can not complete code with high quality (non-compilable), while \ourapproach can complete the code well, and it is \emph{exactly the same} for the reference solution.
    For text-to-code generation, we observe that although both models can not generate exactly the same code as the reference solution, \ourapproach generates a compilable code at the function level.
    These results reveal the effectiveness of our proposed approach for compilable code generation.

	\section{Related Work}
	\paragraph{Neural Code Generation}
	With the rapid development of Deep Learning (DL), some researchers attempt to use DL for code generation tasks. \citet{liu2020a} proposed a neural architecture for code completion task with multi-task learning based on the architecture of Transformer-XL~\citet{Dai2019TransformerXLAL} and BiLSTM~\cite{Schuster1997BidirectionalRN}.
	\citet{kim2021code} presented several ways of feeding the code structure to Transformer~\cite{vaswani2017attention} and further improved the accuracy of the code prediction (next token prediction) task. 
	\citet{Wei2019CodeGA} adopted an encoder-decoder architecture and utilized the relations between code generation and code summarization to improve the performance of both tasks.
	\citet{Yasunaga2021BreakItFixItUL} proposed a new training approach for program repair. They used the critic to check a fixer’s output on real bad inputs and add good outputs to the training data, and trains a breaker to generate realistic bad code from good code.
	\citet{Yasunaga2020GraphbasedSP} used compiler error messages to repair programs. They designed a program-feedback graph and then applied a graph neural network on top to model the reasoning process.
	Many unlabeled programs are used for program repair with self-supervised learning.
	
	Benefiting from the strong power of pre-training techniques~\cite{devlin2019bert, Wang2021ServiceBERTAP} in natural language processing, such as GPT~\cite{radford2018improving}, BART~\cite{lewis2020bart}, and T5~\cite{Raffel2020ExploringTL}, some recent works attempt to pre-train language models on the corpus of source code for code generation. 
	\citet{lu2021codexglue} proposed CodeGPT follows the architecture of GPT-2~\cite{Radford2019LanguageMA}, which is pre-trained with a causal language modeling (CLM) objective on large-scale source code. 
	\citet{ahmad2021unified} proposed PLBART follows the architecture of BART~\cite{lewis2020bart}, which is pre-trained on Java and Python functions paired with code comments via denoising autoencoding. \citet{wang2021codet5} proposed CodeT5 based on the T5~\cite{Raffel2020ExploringTL} architecture, which employs denoising sequence-to-sequence pre-training on multiple programming languages.
	
	\paragraph{Reinforced Text Generation}
	Reinforcement learning~\cite{Sutton1998IntroductionTR} has shown great success in various tasks. It focuses on how agents ought to take actions in an environment to maximize the cumulative reward, is well suited for decision-making tasks. 
	\citet{Ranzato2016SequenceLT} were among the first to apply REINFORCE algorithm~\cite{Williams2004SimpleSG} to train recurrent neural networks on sequence generation tasks, suggesting that directly optimizing the metric used at the test phase can lead to better results.
	\citet{Chen2018FastAS} proposed a hybrid extractive-abstractive architecture with policy-based reinforcement learning. They used an extractor agent to select salient sentences and then employed an abstractor network to rewrite these extracted sentences.
	\citet{wan2018improving,wang2020reinforcement} incorporated the tree structure and sequential content of code snippets and designed a deep reinforcement learning framework optimized by the metric of BLEU to improve the performance of the code summarization task. \citet{Yao2019CoaCorCA} proposed a reinforcement learning framework, which encourages the code annotation model to generate annotations that can be used for code retrieval tasks. \citet{Korbak2021EnergyBasedMF} proposed an energy-based model with an imposed constraint of generating only compilable sequences to improve compilation rates of generated code.
	
	\section{Conclusion and Future Work}
	In this paper, we use the compilability signals in two ways and design a novel method to jointly train the generator and discriminator for compilable code generation, called \ourapproach. 
	Comprehensive experiments on two code generation tasks demonstrate the effectiveness of \ourapproach, improving the average compilation rate of state-of-the-art CodeGPT from 44.18 to 89.18 in the code completion task and from 70.3 to 96.2 in the text-to-code generation task.
	
	
	This work presents our preliminary attempt to generate compilable code. Yet, considering the compilation rate is not the whole story as it still cannot guarantee the correctness of generated code.
	As a future work, we would like to utilize unit tests to evaluate the code correctness towards building more useful code generation models. 
	
	\section*{Acknowledgements}
	This work is supported by National Natural Science Foundation of China under Grant No. 61972290. 
	Yao Wan is partially supported by National Natural Science Foundation of China under Grant No. 62102157. Hao Wu is supported by National Natural Science Foundation of China under Grant No. 61962061.
	\bibliography{ref}
	\bibliographystyle{acl_natbib}

	
	
	
\end{document}